\documentclass{article}

\usepackage{amsmath,amsfonts,bm}

\def\eqref#1{equation~\ref{#1}}

\def\1{\bm{1}}

\DeclareMathAlphabet{\mathsfit}{\encodingdefault}{\sfdefault}{m}{sl}
\SetMathAlphabet{\mathsfit}{bold}{\encodingdefault}{\sfdefault}{bx}{n}

\usepackage{palatino}
\usepackage{amssymb}
\usepackage{amsmath}
\usepackage{amsthm}
\usepackage[a4paper]{geometry}
\usepackage[normalem]{ulem}
\usepackage[round]{natbib}
\usepackage[dvipsnames]{xcolor}
\usepackage[colorlinks,linktoc=true,
             citecolor= Cerulean!85!green!90!black,
             linkcolor=YellowOrange,
             urlcolor=RubineRed!85!black,
             ]{hyperref}
\usepackage{graphicx}

\usepackage{hyperref}
\usepackage{url}
\usepackage{authblk}
\usepackage{makecell}
\usepackage[utf8]{inputenc} 
\usepackage[T1]{fontenc}    
\usepackage{hyperref}       
\usepackage{url}            
\usepackage{booktabs}       
\usepackage{amsfonts}       
\usepackage{nicefrac}       
\usepackage{microtype}      
\usepackage{xcolor}         

\usepackage{graphicx}
\usepackage{subcaption}
\usepackage{framed}
\usepackage{xcolor}
\usepackage{hyperref}
\usepackage{url}
\usepackage{listings}
\usepackage{adjustbox}

\usepackage{enumitem}
\usepackage{microtype}
\usepackage{graphicx}
\usepackage{subcaption}
\usepackage{caption}
\usepackage{wrapfig}

\usepackage{xcolor}
\usepackage{pifont}

\usepackage[utf8]{inputenc} 
\usepackage[T1]{fontenc}    
\usepackage{hyperref}       
\usepackage{url}            
\usepackage{booktabs}       
\usepackage{amsfonts}       
\usepackage{nicefrac}       
\usepackage{microtype}      
\usepackage{xcolor}         

\usepackage{amsmath}
\usepackage{amssymb}
\usepackage{mathtools}
\usepackage{amsthm}
\usepackage{bm}
\usepackage{algorithm}
\usepackage{algpseudocode}
\usepackage{color}
\usepackage{diagbox}
\usepackage{xcolor}
\usepackage{tikz}
\usepackage{tcolorbox}
\tcbuselibrary{breakable}
\theoremstyle{plain}

\theoremstyle{definition}

\theoremstyle{remark}

\usepackage[textsize=tiny]{todonotes}

\usepackage{multirow}



\title{ReCast: Recasting Learning Signals for Reinforcement Learning in Generative Recommendation}

\makeatletter
\newcommand\email[2][]%
   {\newaffiltrue\let\AB@blk@and\AB@pand
      \if\relax#1\relax\def\AB@note{\AB@thenote}\else\def\AB@note{\relax}%
        \setcounter{Maxaffil}{0}\fi
      \begingroup
        \let\protect\@unexpandable@protect
        \def\thanks{\protect\thanks}\def\footnote{\protect\footnotetext{\textdagger\ #1}}%
        \@temptokena=\expandafter{\AB@authors}%
        {\def\\{\protect\\\protect\Affilfont}\xdef\AB@temp{#2}}%
         \xdef\AB@authors{\the\@temptokena\AB@las\AB@au@str
         \protect\\[\affilsep]\protect\Affilfont\AB@temp}%
         \gdef\AB@las{}\gdef\AB@au@str{}%
        {\def\\{, \ignorespaces}\xdef\AB@temp{#2}}%
        \@temptokena=\expandafter{\AB@affillist}%
        \xdef\AB@affillist{\the\@temptokena \AB@affilsep
          \AB@affilnote{}\protect\Affilfont\AB@temp}%
      \endgroup
       \let\AB@affilsep\AB@affilsepx
}
\makeatother

\author[1]{Peiyan Zhang}
\author[1]{Hanmo Liu}
\author[1]{Chengxuan Tong}
\author[1]{Yuxia Wu}
\author[1]{Wei Guo}
\author[1]{Yong Liu}

\affil[1]{Huawei Technologies Co., Ltd.}
\email{\texttt{\{zhangpeiyan2, wuyuxia6\}@h-partners.com}}
\email{\texttt{\{liu.hanmo, tong.chengxuan, guowei67, liu.yong6\}@huawei.com}}

\date{}

\begin{document}
\maketitle

\begin{abstract}

Generic group-based RL assumes that sampled rollout groups are already usable learning signals. We show that this assumption breaks down in sparse-hit generative recommendation, where many sampled groups never become learnable at all. We propose ReCast, a repair-then-contrast learning-signal framework that first restores minimal learnability for all-zero groups and then replaces full-group reward normalization with a boundary-focused contrastive update on the strongest positive and the hardest negative. ReCast leaves the outer RL framework unchanged, modifies only within-group signal construction, and partially decouples rollout search width from actor-side update width.

Across multiple generative recommendation tasks, ReCast consistently outperforms OpenOneRec-RL, achieving up to 36.6\% relative improvement in Pass@1. Its matched-budget advantage is substantially larger: ReCast reaches the baseline's target performance with only 4.1\% of the rollout budget, and this advantage widens with model scale. The same design also yields direct system-level gains, reducing actor-side update time by 16.60$\times$, lowering peak allocated memory by 16.5\%, and improving actor MFU by 14.2\%. Mechanism analysis shows that ReCast mitigates the persistent all-zero / single-hit regime, restores learnability when natural positives are scarce, and converts otherwise wasted rollout budget into more stable policy updates. These results suggest that, for generative recommendation, the decisive RL problem is not only how to assign rewards, but how to construct learnable optimization events from sparse, structured supervision.


\end{abstract}

\section{Introduction}
\label{sec:intro}

\begin{figure*}[h]
    \centering
    \begin{subfigure}[t]{0.48\textwidth}
        \centering
        \includegraphics[width=\linewidth]{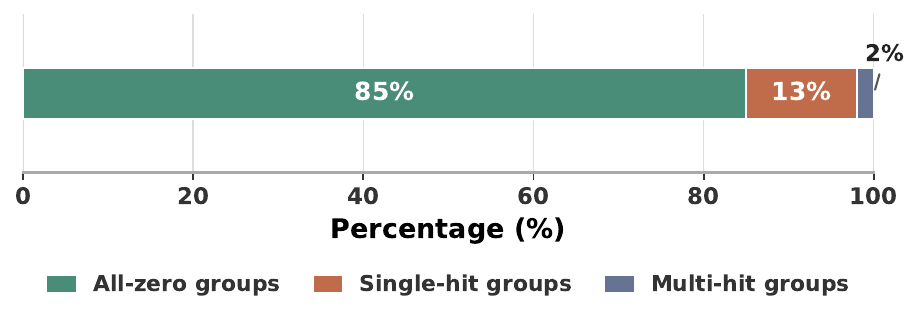}
        \caption{Group composition.}
        \label{fig:intro_group_composition}
    \end{subfigure}
    \hfill
    \begin{subfigure}[t]{0.48\textwidth}
        \centering
        \includegraphics[width=\linewidth]{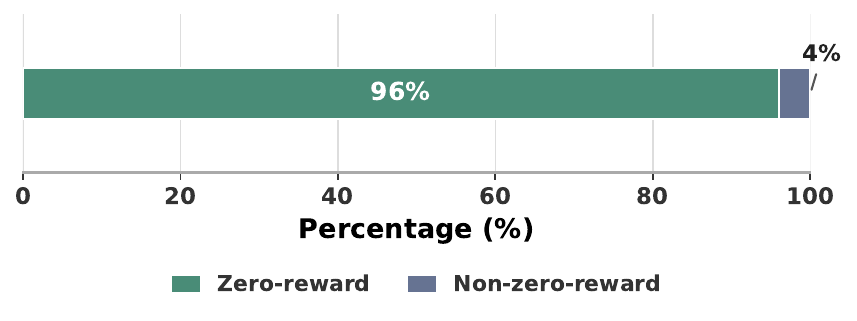}
        \caption{Sample reward composition.}
        \label{fig:intro_sample_reward_composition}
    \end{subfigure}
    \caption{Signal degeneracy in the RL stage of OpenOneRec under a representative sparse-hit setting.}
    \label{fig:grpo_degenerate_intro}
\end{figure*}

Generative recommendation replaces candidate scoring with direct item generation \citep{p5,cui2022m6recgenerativepretrainedlanguage,tallrec,llara,rajput2023recommender,10597986,letter,eager,eagerllm,unger,colarec}, and RL-style post-training is increasingly adopted to optimize hit-oriented metrics such as Pass@K \citep{wang2025gflowgr}. Existing methods largely inherit generic group-based RL \citep{jiang2026spendsearchpaysvalueguided,zhang2026reinforced}: sample a group of outputs, score them, and update the policy from group-level rewards~\citep{lin2026recr1bridginggenerativelarge, zhou2025openonerec}. This assumes that a sampled and scored group already constitutes a usable learning unit. Under sparse-hit recommendation, that assumption often fails. The issue is not reward sparsity alone, but that many sampled groups never become trainable learning units in the first place.

This mismatch is most severe in sparse-hit, single-target recommendation, where natural positives are rare and three failure modes dominate. All-zero groups are unlearnable because they contain no positive-negative boundary. Single-hit groups are trainable but fragile, with updates dominated by one accidental hit and noisy group statistics. Binary supervision further collapses structured near misses into the same zero-reward class as fully irrelevant outputs \citep{ding2026towards, xie2026sagesequenceleveladaptivegradient}. Figure~\ref{fig:grpo_degenerate_intro} shows that this is not a corner case. In a representative sparse-hit OpenOneRec~\citep{zhou2025openonerec} RL setting, about 85\% of sampled groups remain all-zero after 20K steps, another 13\% are single-hit, and only 2\% exhibit richer reward structure; at the sample level, about 96\% of responses still receive zero reward. Much of the rollout budget therefore remains trapped in unlearnable or weakly learnable groups rather than producing reliable policy improvement.

Building on these observations, we propose \textbf{ReCast}, a \textbf{repair-then-contrast} learning-signal framework that modifies only within-group signal construction in generative recommendation. ReCast addresses two sequential questions: whether a sampled group is learnable at all, and how it should be updated once learning becomes possible. It first performs rollout repair, which restores minimal learnability for all-zero groups by injecting a valid positive anchor. It then applies a boundary contrastive update that updates only the strongest positive and the hardest negative, replacing full-group reward normalization with a local decision-boundary update. The rollout process and outer RL objective remain unchanged.

ReCast also changes the update path of group-based RL. Conventional methods~\citep{zhou2025onerec,liu2025onerecthinkintextreasoninggenerative} backpropagate through the full sampled group, so actor-side update cost grows with group width. ReCast instead updates only a constant-size active subset defined by the local positive-negative boundary. It therefore preserves rollout search width while avoiding full-group actor-side updates, partially decoupling search from update. This makes ReCast a lightweight drop-in replacement whose advantage becomes more pronounced at larger scales and under sparser-hit regimes.

Experiments show that ReCast improves not only final recommendation quality, but also the scaling efficiency of RL post-training. Under the same rollout budget, ReCast consistently outperforms the baseline across multiple generative recommendation tasks, with up to 36.6\% relative improvement in Pass@1. Its matched-budget advantage is substantially larger: ReCast reaches the baseline's target performance with only 4.1\% of the rollout budget. This advantage widens with model scale and search width, and materializes directly in system measurements: ReCast reduces actor-side update time by 16.60$\times$, lowers peak allocated memory by 16.5\%, and improves actor MFU by 14.2\%.

\paragraph{Contributions.}
\begin{itemize}[leftmargin=*,noitemsep,topsep=0pt]
    \item \textbf{Learnability restoration for sparse-hit recommendation RL.} We identify a mismatch between generic group-based RL and sparse-hit generative recommendation: sampled groups are easy to obtain, but many never become usable learning units. We address this mismatch with \textbf{ReCast}, a repair-then-contrast signal design that restores minimal learnability for all-zero groups and refines the most informative local positive--negative boundary once learning becomes possible.

    \item \textbf{Constant-size boundary-focused updating.} ReCast leaves the rollout process and outer RL objective unchanged, but reduces actor-side update width from \(\mathbf{O(G)}\) to \(\mathbf{O(1)}\), where \(G\) is the rollout group size, by updating only a constant-size subset. This partially decouples rollout search width from actor-side optimization cost, turning wider search into a more favorable scaling knob for recommendation RL.

    \item \textbf{Comprehensive evaluation and analysis.} Across multiple generative recommendation tasks, ReCast improves final recommendation quality, enters the useful-learning regime substantially earlier than the baseline, and shows widening matched-budget advantages with model scale. System measurements further show that the constant-size update design translates into lower actor-side cost, faster training, reduced memory footprint, and higher hardware utilization. 
\end{itemize}

\section{Motivation}
\label{sec:motivation}

\noindent\textbf{Notation.}
We consider offline generative recommendation in the single-target next-item setting. Given a prompt $q \in \mathcal{Q}$, the policy generates a response
\[
R=(y_1,\ldots,y_T)\sim \pi_\theta(\cdot\mid q),
\]
and the ground-truth target item is denoted by $s\in\mathcal{S}$. RL is applied on top of a post-SFT backbone to optimize hit-oriented recommendation metrics.

\subsection{Default Group-Based RL Assumes Learnable Groups}

A common default view of recommendation RL is to optimize over sampled groups. For each prompt $q$, a rollout policy $\pi_{\text{old}}$ samples a group of $G$ candidate responses
\[
\mathcal{G}(q)=\{R_1,\ldots,R_G\},
\qquad
R_i\sim \pi_{\text{old}}(\cdot\mid q).
\]
Each response $R_i$ receives a scalar reward
\[
r_i=r(R_i;q,s),
\]
which measures its quality with respect to the target item $s$.

In the single-target setting studied here, we focus on sparse binary hit rewards, i.e., $r_i\in\{0,1\}$, where $r_i=1$ indicates that the target item $s$ is successfully generated. A representative default choice is a GRPO-style group-based update \citep{shao2024deepseekmath,zheng2025groupsequencepolicyoptimization}, which constructs within-group advantages by reward normalization:
\[
\mu_q=\frac{1}{G}\sum_{i=1}^{G}r_i,
\qquad
\sigma_q=\sqrt{\frac{1}{G}\sum_{i=1}^{G}(r_i-\mu_q)^2},
\qquad
\hat{A}_i=\frac{r_i-\mu_q}{\sigma_q+\epsilon},
\]
where $\epsilon>0$ is a small constant for numerical stability. The resulting policy objective is
\[
\mathcal{L}_{\text{GRPO}}(\theta)
=
\mathbb{E}_{q\sim\mathcal{D},\,\mathcal{G}(q)\sim\pi_{\text{old}}}
\left[
\frac{1}{G}\sum_{i=1}^{G}
\hat{A}_i \log \pi_\theta(R_i\mid q)
\right]
-\beta\,\mathrm{KL}\!\left(
\pi_\theta(\cdot\mid q)\,\|\,\pi_{\mathrm{ref}}(\cdot\mid q)
\right),
\]
where $\pi_{\mathrm{ref}}$ is a reference policy and $\beta>0$ controls KL regularization.

The key assumption behind this default view is that a sampled and scored group already constitutes a usable learning unit.

\subsection{Sparse-Hit Supervision Breaks This Assumption}

We focus on sparse binary hit rewards, where $r_i\in\{0,1\}$ and $r_i=1$ indicates a successful hit on the target item. Let
\[
K(q)=\sum_{i=1}^{G} r_i
\]
denote the number of hits in a sampled group. Under sparse-hit supervision, three failure modes dominate.

\begin{itemize}[leftmargin=*,noitemsep,topsep=0pt]
\item \textbf{All-zero groups.}
When $K(q)=0$, the sampled group contains no positive--negative boundary at all. All responses receive zero reward, so the group is unlearnable as a policy-improvement unit.

\item \textbf{Single-hit groups.}
When $K(q)=1$, learning becomes possible but fragile. The update is determined by one accidental positive together with group statistics, making it highly sensitive to sampling noise and group composition.

\item \textbf{Near-miss collapse.}
Binary rewards flatten structurally or semantically close responses into the same zero-reward class as fully irrelevant negatives. As a result, potentially informative recommendation structure is discarded rather than used to refine the decision boundary.
\end{itemize}

These failures show that, under sparse-hit supervision, group rollout does not automatically yield a reliable learning signal. What becomes scarce is not rollout budget alone, but usable learning units.

\subsection{Design Requirements}

These failure modes suggest that the central design problem is not only how to optimize a sampled signal, but how to construct one that is learnable first and informative next. We have three requirements follow. 
\begin{enumerate}[leftmargin=*,noitemsep,topsep=0pt]
    \item Learning should not depend on lucky hits: all-zero groups should be recoverable rather than wasted.
    \item Once a group becomes learnable, the update should focus on the local decision boundary rather than coarse group statistics.
    \item Recommendation structure should be used rather than collapsed, so that the target and structured near misses can help define more informative within-group contrasts.
\end{enumerate}

This directly motivates a two-step signal design: restore learnability first, then refine the boundary.

\section{ReCast: Repair-then-Contrast Signal Design}
\label{sec:method}

\begin{figure}[t]
    \centering
    \includegraphics[width=0.95\linewidth]{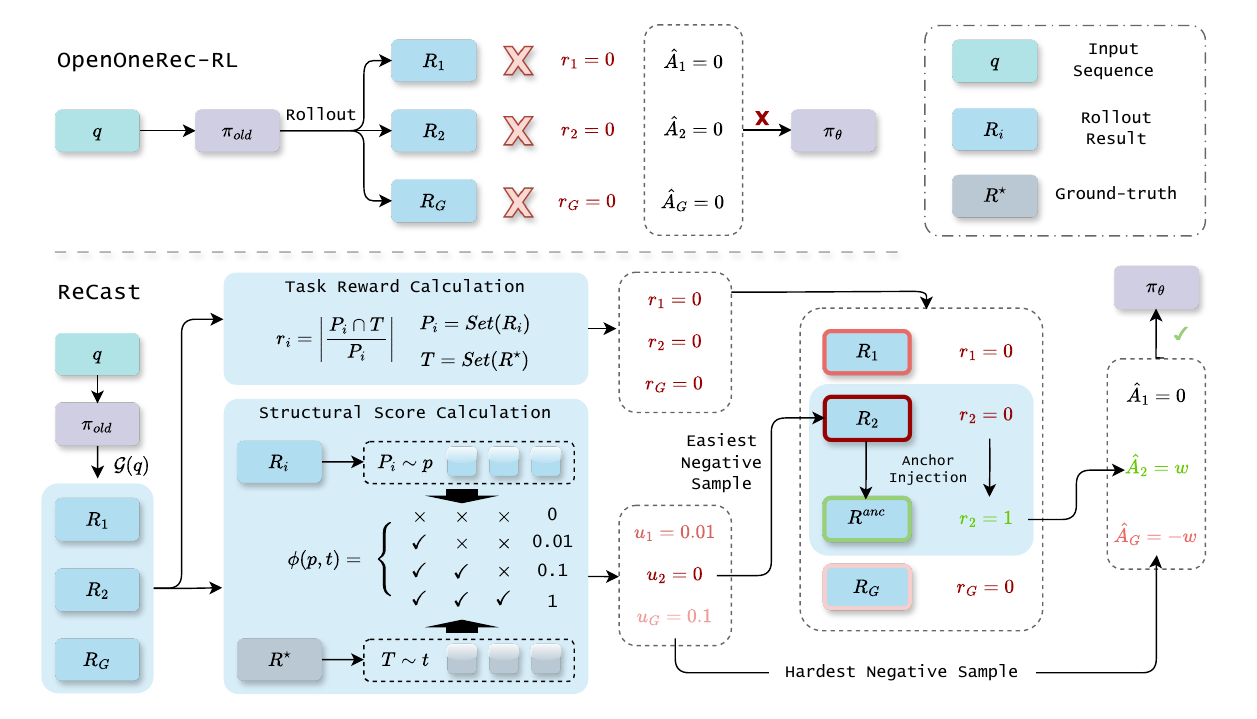}
   \caption{Comparison between \textbf{OpenOneRec-RL} and \textbf{ReCast}. OpenOneRec-RL updates from group-relative reward normalization over the sampled rollout group, whereas ReCast first repairs all-zero groups to restore minimal learnability and then applies a boundary-focused update on the strongest positive and the hardest negative. The outer RL framework remains unchanged.}
    \label{fig:recast_overview}
\end{figure}

ReCast leaves rollout sampling and the outer RL objective unchanged, and modifies only within-group signal construction. The method has two steps. It first restores minimal learnability for all-zero groups through rollout repair, and then refines the strongest local positive--negative boundary through a constant-size contrastive update. Figure~\ref{fig:recast_overview} summarizes the full signal path.

\subsection{Task Reward and Structural Score}

For a sampled response $R_i$ and ground-truth output $R^\star$, let
\[
P_i=\operatorname{Set}(R_i), \qquad T=\operatorname{Set}(R^\star)
\]
denote the extracted target-ID sets from the prediction and the ground truth, respectively.
Although we study single-target next-item recommendation, a generated response may still contain zero, one, or multiple valid IDs after parsing, so we write both signals in set form.

\paragraph{Task reward.}
We define the task reward as
\[
r_i=
\begin{cases}
0, & \text{if } P_i=\varnothing \text{ or } T=\varnothing,\\[4pt]
\dfrac{|P_i\cap T|}{|P_i|}, & \text{otherwise}.
\end{cases}
\]
In the single-target single-prediction case, this reduces to the standard binary hit reward.

\paragraph{Structural score.}
To expose recommendation structure beyond exact hits, we define a structural score used only for within-group signal construction. For two target IDs
\[
p=(p_a,p_b,p_c), \qquad t=(t_a,t_b,t_c),
\]
let
\[
\phi(p,t)=
\begin{cases}
1, & \text{if } p=t,\\
0.1, & \text{if } (p_a,p_b)=(t_a,t_b)\text{ and }p_c\neq t_c,\\
0.01, & \text{if } p_a=t_a \text{ and } p_b\neq t_b,\\
0, & \text{otherwise}.
\end{cases}
\]
The structural score of response $R_i$ is then
\[
u_i=
\begin{cases}
0, & \text{if } P_i=\varnothing \text{ or } T=\varnothing,\\[4pt]
\dfrac{1}{|P_i|}\sum_{p\in P_i}\max_{t\in T}\phi(p,t), & \text{otherwise}.
\end{cases}
\]

The key distinction is that $r_i$ defines task success, whereas $u_i$ is used only to rank structural proximity within the sampled group. Thus, ReCast leaves the outer reward unchanged and uses $u_i$ only for within-group signal construction.

\subsection{Rollout Repair}

Rollout repair is invoked only when a sampled group contains no positive at all. Formally, given
\[
\mathcal{G}(q)=\{R_1,\ldots,R_G\},
\]
repair is applied only if
\[
\sum_{i=1}^{G} r_i = 0.
\]
In that case, ReCast constructs a valid positive anchor response $R^{\mathrm{anc}}$ such that
\[
r(R^{\mathrm{anc}};q,s) > 0.
\]
In practice, $R^{\mathrm{anc}}$ can be derived from the ground-truth output $R^\star$. This restores minimal learnability without changing the outer objective.

To keep the intervention minimal, ReCast replaces the least informative response in the group:
\[
j_{\mathrm{rep}}=\arg\min_{1\le i\le G} u_i,
\]
yielding the repaired group
\[
\widetilde{\mathcal{G}}(q)=\bigl(\mathcal{G}(q)\setminus \{R_{j_{\mathrm{rep}}}\}\bigr)\cup\{R^{\mathrm{anc}}\}.
\]
Since $u_i$ ranks structural proximity to the target, this replacement preserves the more informative negatives whenever possible.

After repair, ReCast recomputes $\{r_i,u_i\}$ on $\widetilde{\mathcal{G}}(q)$ and proceeds to the boundary update step. If the sampled group already contains at least one positive, no repair is applied.

\subsection{Boundary Contrastive Update}

Given a repaired or already trainable group $\widetilde{\mathcal{G}}(q)$, ReCast updates only a local positive--negative boundary. We select
\[
i^{+}=\arg\max_{i:\, r_i>0} r_i,
\qquad
i^{-}=\arg\max_{i:\, r_i=0} u_i.
\]
Thus, the positive is chosen by task success, while the negative is chosen by structural proximity to the target.

ReCast then assigns learning signal only to this pair:
\begin{equation}
A_i^{\mathrm{ReCast}}=
\begin{cases}
+w, & i=i^{+},\\
-w, & i=i^{-},\\
0, & \text{otherwise},
\end{cases}
\label{eq:recast_adv}
\end{equation}
where $w>0$ is a fixed contrastive weight (default $w=1$). This replaces full-group reward normalization with a constant-size active subset update. Once the boundary pair has been selected, the objective is to move probability mass from the hardest non-hit toward the strongest positive rather than to redistribute credit across the full group.\footnote{If no valid non-hit response exists in an otherwise trainable group, we skip the contrastive update for definitional completeness. This case is negligible in our sparse-hit settings.}

\subsection{Outer Objective}

ReCast leaves the outer RL framework unchanged and modifies only how within-group advantages are constructed. For each prompt $q$, a rollout group $\mathcal{G}(q)$ is first sampled from $\pi_{\mathrm{old}}$. If the group is all-zero, rollout repair produces a trainable group $\widetilde{\mathcal{G}}(q)$; otherwise, we let $\widetilde{\mathcal{G}}(q)=\mathcal{G}(q)$. The ReCast advantage vector $A_i^{\mathrm{ReCast}}$ is then computed on $\widetilde{\mathcal{G}}(q)$ by the boundary contrastive update in Eq.~(\ref{eq:recast_adv}).

The resulting objective is
\[
\mathcal{L}_{\mathrm{ReCast}}(\theta)
=
\mathbb{E}_{q\sim\mathcal{D},\,\mathcal{G}(q)\sim\pi_{\mathrm{old}}}
\left[
\frac{1}{G}\sum_{i=1}^{G}
A_i^{\mathrm{ReCast}} \log \pi_\theta(R_i\mid q)
\right]
-
\beta\,\mathrm{KL}\!\left(\pi_\theta(\cdot\mid q)\,\|\,\pi_{\mathrm{ref}}(\cdot\mid q)\right),
\]
where $\pi_{\mathrm{ref}}$ is a reference policy and $\beta>0$ controls KL regularization. Thus, ReCast changes only the within-group signal path before the standard KL-regularized policy update.

\section{Search--Update Decoupling}
\label{sec:scaling}

Conventional group-based RL ties rollout width and actor-side update width to the same variable \(G\). Under sparse-hit generative recommendation, this is inefficient: broader rollout is useful to expose rare positives, but actor learning need not scale with the full sampled group. ReCast breaks this coupling by keeping search wide and updates local.

\subsection{From Coupled Group Width to Boundary-Supported Updates}

In conventional group-based RL, the group size \(G\) controls two quantities at once. It sets the search width by determining how many candidates are sampled per prompt, and it sets the update width by determining how many samples enter actor-side forward--backward optimization. Under sparse-hit recommendation, these two roles favor different operating points: search benefits from larger \(G\), since natural positives are rare, whereas actor-side learning need only depend on the small subset that defines the local decision boundary.

This difference appears directly at the level of update support. Under full-group updating, all sampled responses support actor learning:
\[
W_{\mathrm{search}} = G,
\qquad
W_{\mathrm{update}}^{\mathrm{base}} = G.
\]
ReCast changes only the second quantity. Repair restores minimal learnability when needed, and boundary contrast reduces actor-side learning to one positive and one negative:
\[
W_{\mathrm{search}} = G,
\qquad
W_{\mathrm{update}}^{\mathrm{ReCast}} = O(1).
\]

The sampled group therefore changes role. Under conventional group-based RL, it is both the search object and the full actor-side optimization burden. Under ReCast, it is primarily a search pool: rollout remains broad enough to expose rare positives and informative negatives, while actor-side learning depends only on the extracted local boundary.

Let \(c_{\mathrm{roll}}\) denote the rollout cost of one sampled response and \(c_{\mathrm{upd}}\) the actor-side update cost of one response. The per-group cost is then
\[
C_{\mathrm{base}} = G\,c_{\mathrm{roll}} + G\,c_{\mathrm{upd}},
\qquad
C_{\mathrm{ReCast}} = G\,c_{\mathrm{roll}} + O(1)\,c_{\mathrm{upd}}.
\]
ReCast thus preserves search width while removing the linear dependence of actor-side learning on \(G\).

\subsection{Implications for Scaling}

This decoupling changes the scaling regime of group-based RL.

\begin{itemize}[leftmargin=*,noitemsep,topsep=2pt]
    \item \textbf{Model scale.} The advantage of ReCast should grow with model size, since actor-side optimization becomes more expensive for larger backbones while update support remains constant.

    \item \textbf{Search width.} Increasing \(G\) should be more favorable under ReCast. Larger rollout still improves candidate coverage, but no longer expands actor-side learning cost in proportion to group width.

    \item \textbf{System efficiency.} By keeping actor-side learning on a constant-size active subset, ReCast directly reduce wall-clock time, memory footprint, and actor-side workload relative to full-group updating.
\end{itemize}

\section{Experiments}
\label{sec:eval}

We evaluate ReCast through the following three questions:

\begin{itemize}[leftmargin=*,noitemsep,topsep=2pt]
    \item \textbf{Q1:} Under the same training budget, does ReCast improve final recommendation quality and enter the useful-learning regime earlier than baseline methods?

    \item \textbf{Q2:} Why does ReCast help? In particular, does it mitigate the persistent degenerate signal regime in the baseline, and what roles do repair and boundary-focused updating play?

    \item \textbf{Q3:} Does the search--update decoupling view translate into scaling behavior and system-level efficiency, \textit{e.g.,} scaling with model size, scaling with rollout width, and lower training cost?
\end{itemize}

\subsection{Experimental Setup}

\paragraph{Benchmarks.}
We evaluate ReCast under the OpenOneRec benchmark setting on RecIF-Bench \citep{zhou2025openonerec}, covering five generative recommendation tasks: \textbf{Ad Recommendation}, \textbf{Product Recommendation}, \textbf{Short Video Recommendation}, \textbf{Interactive Recommendation}, and \textbf{Label-Conditional Recommendation}. We follow the original task definitions, data preprocessing, and evaluation protocol.

\paragraph{Controlled comparison.}
All methods are built on the same OpenOneRec generative recommendation pipeline and initialized from the same \textbf{post-SFT} checkpoint. Unless otherwise specified, all RL variants share the same rollout temperature, group size, batch size, optimizer settings, KL coefficient, maximum generation length, and total RL budget. Our main baseline is the original \textbf{OpenOneRec-RL} pipeline \citep{zhou2025openonerec}, whose RL stage applies GRPO-style within-group reward normalization. The comparison therefore isolates the effect of RL-stage signal construction.

\paragraph{Backbone scales.}
Unless otherwise specified, all experiments use \textbf{Qwen3-1.7B}. The model-scaling study in Section~\ref{sec:scale_model_size} uses \textbf{Qwen3-1.7B}, \textbf{Qwen3-8B}, and \textbf{Qwen3-14B}. The search-width scaling study in Section~\ref{sec:scale_search_width} and the system-efficiency results in Section~\ref{sec:efficiency} use \textbf{Qwen3-8B}, where the effect of actor-side workload reduction is more clearly exposed.

\paragraph{Execution and implementation.}
Training is conducted on an Ascend/NPU cluster with 64 NPUs in total. All compared methods use the same hardware configuration within each controlled experiment. For ReCast runs, inactive samples are physically filtered before actor-side \texttt{old\_log\_prob}, \texttt{ref\_log\_prob}, and \texttt{update\_actor}, so the reported system-efficiency gains reflect realized workload reduction rather than post-hoc masking.

\paragraph{Metrics.}
We report final offline recommendation quality using \textbf{Pass@1}, \textbf{Pass@32}, and \textbf{Recall@32}. To evaluate learning efficiency and scaling behavior, we additionally report matched-budget comparisons and scaling and system-efficiency results under the same comparison protocol.

\begin{table*}[h]
\centering
\small
\setlength{\tabcolsep}{5pt}
\renewcommand{\arraystretch}{1.05}
\caption{Main results on RecIF-Bench.}
\label{tab:main_results}
\begin{tabular}{llccc}
\toprule
\textbf{Task} & \textbf{Metric} & \textbf{OpenOneRec-RL} & \textbf{ReCast} & \textbf{Rel. Improv.} \\
\midrule
\multirow{3}{*}{Short Video Rec}
& Pass@1    & 0.0285 & \textbf{0.0311} & \textbf{9.12\%$\uparrow$} \\
& Pass@32   & 0.1088 & \textbf{0.1140} & \textbf{4.78\%$\uparrow$} \\
& Recall@32 & 0.0150 & \textbf{0.0158} & \textbf{5.33\%$\uparrow$} \\
\midrule
\multirow{3}{*}{Ad Rec}
& Pass@1    & 0.0138 & \textbf{0.0160} & \textbf{15.90\%$\uparrow$} \\
& Pass@32   & 0.1689 & \textbf{0.1782} & \textbf{5.50\%$\uparrow$} \\
& Recall@32 & 0.0558 & \textbf{0.0602} & \textbf{7.90\%$\uparrow$} \\
\midrule
\multirow{3}{*}{Product Rec}
& Pass@1    & 0.0230 & \textbf{0.0252} & \textbf{9.60\%$\uparrow$} \\
& Pass@32   & 0.1694 & \textbf{0.1882} & \textbf{11.10\%$\uparrow$} \\
& Recall@32 & 0.0420 & \textbf{0.0465} & \textbf{10.70\%$\uparrow$} \\
\midrule
\multirow{3}{*}{Label-Cond.\ Rec}
& Pass@1    & 0.0041 & \textbf{0.0056} & \textbf{36.60\%$\uparrow$} \\
& Pass@32   & 0.0289 & \textbf{0.0292} & \textbf{1.00\%$\uparrow$} \\
& Recall@32 & 0.0116 & \textbf{0.0120} & \textbf{3.40\%$\uparrow$} \\
\midrule
\multirow{3}{*}{Interactive Rec}
& Pass@1    & 0.0860 & \textbf{0.0970} & \textbf{12.80\%$\uparrow$} \\
& Pass@32   & 0.4310 & \textbf{0.4390} & \textbf{1.86\%$\uparrow$} \\
& Recall@32 & 0.2865 & \textbf{0.2973} & \textbf{3.80\%$\uparrow$} \\
\bottomrule
\end{tabular}
\end{table*}

\subsection{Main Results (for Q1)}
\label{sec:main}

Table~\ref{tab:main_results} shows that ReCast consistently outperforms OpenOneRec-RL across all five generative recommendation tasks under the same RL budget. The largest gains appear on \textbf{Pass@1}, where ReCast improves first-hit quality by \textbf{9.1\%} to \textbf{36.6\%}. The gains also extend to \textbf{Pass@32} and \textbf{Recall@32}, indicating that ReCast improves not only the top-ranked hit, but also the overall accessibility of the target item within generated outputs.

\begin{figure*}[h]
    \centering
    \begin{subfigure}[t]{0.32\textwidth}
        \centering
        \includegraphics[width=\linewidth]{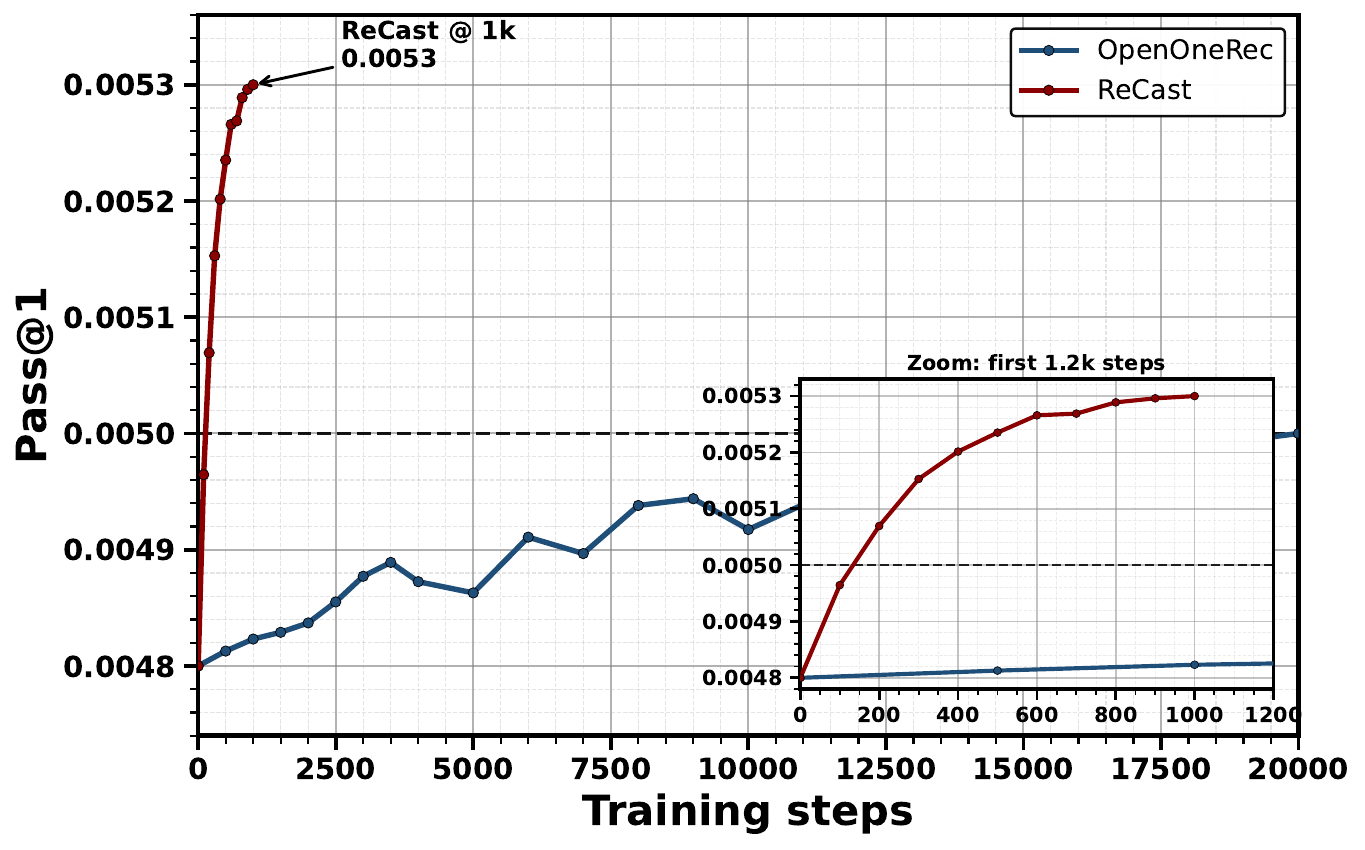}
        \caption{Label-Cond. Recommendation.}
        \label{fig:early_stage_label}
    \end{subfigure}
    \hfill
    \begin{subfigure}[t]{0.32\textwidth}
        \centering
        \includegraphics[width=\linewidth]{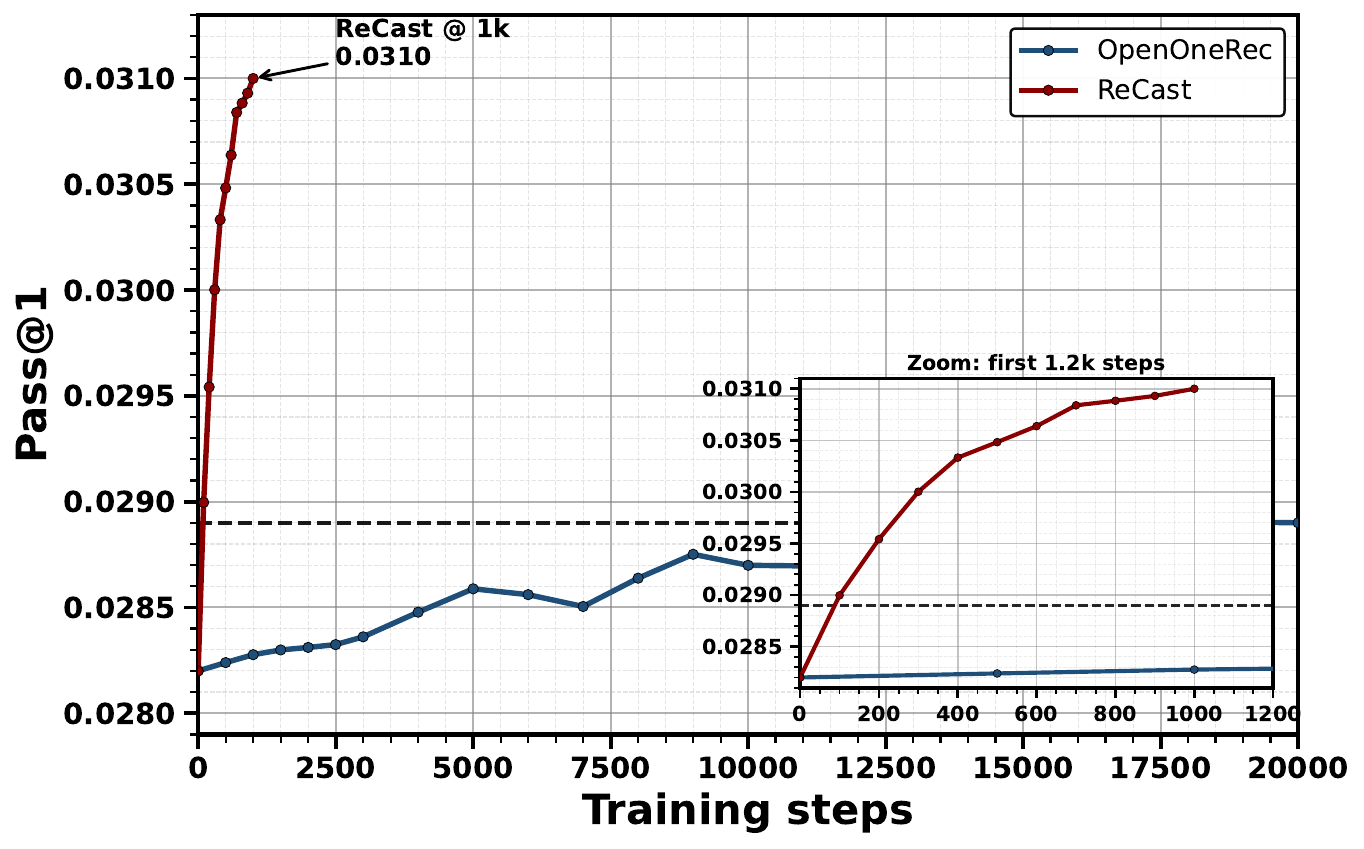}
        \caption{Short Video Recommendation.}
        \label{fig:early_stage_video}
    \end{subfigure}
    \hfill
    \begin{subfigure}[t]{0.32\textwidth}
        \centering
        \includegraphics[width=\linewidth]{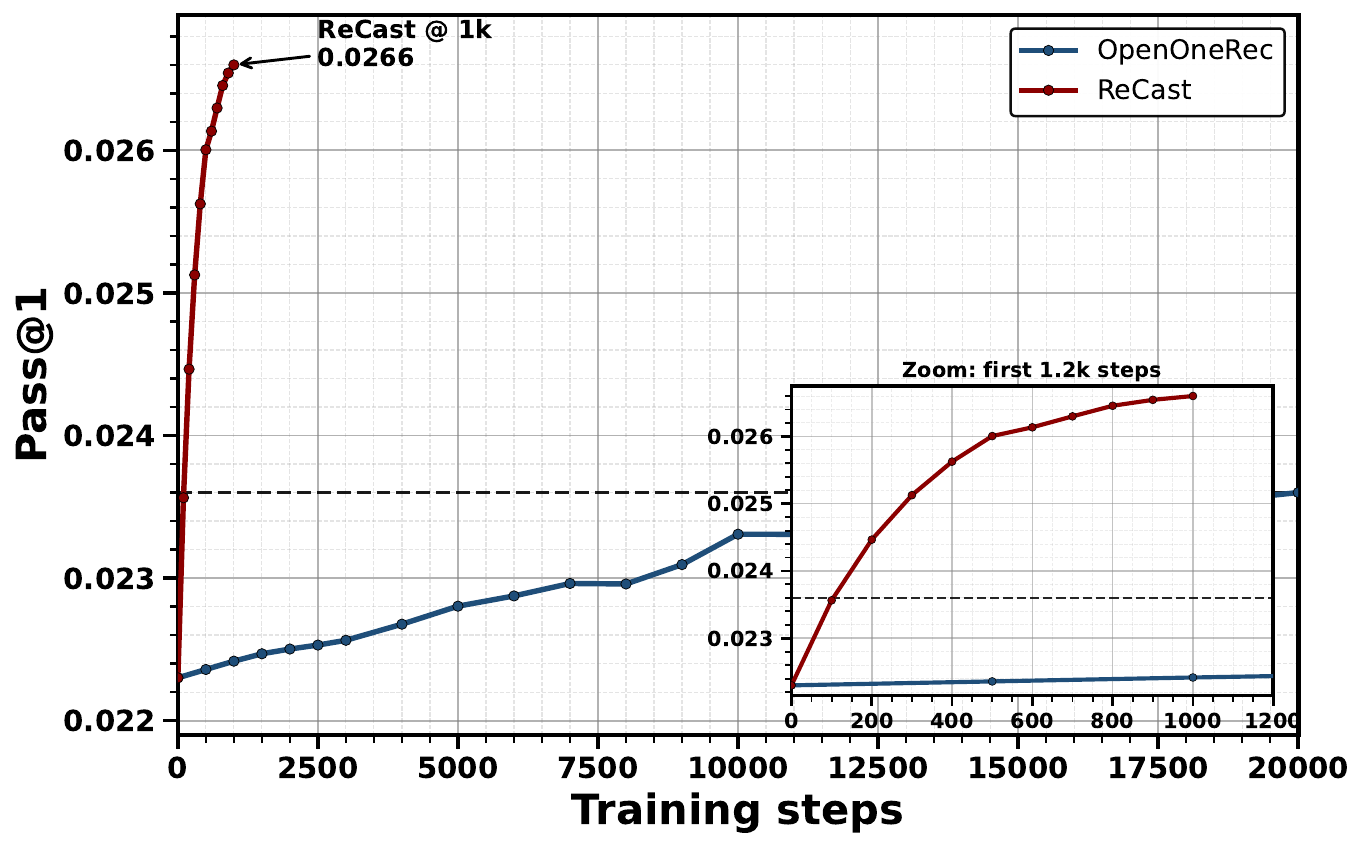}
        \caption{Product Recommendation.}
        \label{fig:early_stage_product}
    \end{subfigure}
    \caption{Early-stage learning efficiency on three representative tasks.}
    \label{fig:early_stage_efficiency}
\end{figure*}

The advantage is not limited to final quality. Figure~\ref{fig:early_stage_efficiency} shows that ReCast enters the useful-learning regime substantially earlier. Across the evaluated tasks, \textbf{ReCast at 1K steps already surpasses OpenOneRec-RL at 20K steps}. On \textbf{Label-Conditional Recommendation}, for example, ReCast reaches \textbf{Pass@1 = 0.0053} at \textbf{1K} steps, exceeding the \textbf{0.0050} achieved by OpenOneRec-RL at \textbf{20K} steps. Since \textbf{20K} steps still lie well within the first epoch in our setting, this gap reflects genuinely earlier useful learning rather than late-stage degradation of the baseline.

\subsection{Mechanism Analysis (for Q2)}

The previous results show that ReCast improves both final quality and early learning efficiency. We now ask why. The analysis below isolates four pieces of the mechanism: persistent signal degeneracy in the baseline, learnability restoration by repair, optimization stability from boundary-focused updating, and the regime-dependent roles of the two components.

\subsubsection{Persistent Signal Degeneracy}

Table~\ref{tab:signal_degeneracy} shows that under sparse-hit supervision, OpenOneRec-RL does not escape the all-zero / single-hit regime after a brief early phase. The same pattern persists throughout training. From 1K to 40K steps, the all-zero group ratio changes only from \textbf{88\%} to \textbf{85\%}, the single-hit group ratio from \textbf{10\%} to \textbf{13\%}, and the zero-reward sample ratio from \textbf{97.2\%} to \textbf{96.0\%}. Much of the rollout budget therefore still fails to become reliable policy-improvement signal. The bottleneck is not only sparse reward at the sample level, but persistent degeneracy at the group level: many sampled groups never become informative learning units in the first place.

\begin{table}[h]
\centering
\small
\setlength{\tabcolsep}{4.8pt}
\renewcommand{\arraystretch}{1.08}
\caption{Persistent signal degeneracy under OpenOneRec-RL. }
\label{tab:signal_degeneracy}
\begin{tabular}{lcccc}
\toprule
\textbf{Metric} & \textbf{1K} & \textbf{20K} & \textbf{40K} & \textbf{$\Delta$(40K--1K)} \\
\midrule
All-zero group ratio      & 88\%   & 85\%   & 85\%   & -3\% \\
Single-hit group ratio    & 10\%   & 13\%   & 13\%   & +3\% \\
Zero-reward sample ratio  & 97.2\% & 96.3\% & 96.0\% & -1.2\% \\
\bottomrule
\end{tabular}
\end{table}

\subsubsection{Repair Restores Learnability}

Table~\ref{tab:repair_dynamics} summarizes the role of repair. It first restores learnability immediately, lifting the valid trainable-group ratio from \textbf{13\%} to \textbf{100\%}. It then becomes progressively less necessary: the repair trigger ratio declines from \textbf{88\%} at 1K steps to \textbf{61\%} at 10K and \textbf{44\%} at 20K, while the naturally trainable-group ratio rises from \textbf{12\%} to \textbf{39\%} and \textbf{56\%}, respectively. Repair therefore matters most when natural positives are scarce, and recedes once the policy begins to produce trainable groups on its own.

\begin{table}[h]
\centering
\small
\setlength{\tabcolsep}{4.5pt}
\renewcommand{\arraystretch}{1.08}
\caption{Repair first restores learnability, and then becomes less necessary as training progresses.}
\label{tab:repair_dynamics}
\begin{minipage}[t]{0.38\linewidth}
\centering
\textbf{Immediate effect of repair}

\vspace{4pt}
\begin{tabular}{lc}
\toprule
\textbf{Metric} & \textbf{Value} \\
\midrule
Before repair & 13\% \\
After repair  & 100\% \\
\bottomrule
\end{tabular}
\end{minipage}
\hfill
\begin{minipage}[t]{0.58\linewidth}
\centering
\textbf{Repair recedes over training}

\vspace{4pt}
\begin{tabular}{lccc}
\toprule
\textbf{Metric} & \textbf{1K} & \textbf{10K} & \textbf{20K} \\
\midrule
Repair trigger ratio           & 88\% & 61\% & 44\% \\
Naturally trainable-group ratio & 12\% & 39\% & 56\% \\
\bottomrule
\end{tabular}
\end{minipage}
\end{table}

\subsubsection{Boundary-Focused Stability}

\begin{figure*}[h]
    \centering
    \begin{subfigure}[t]{0.32\textwidth}
        \centering
        \includegraphics[width=\linewidth]{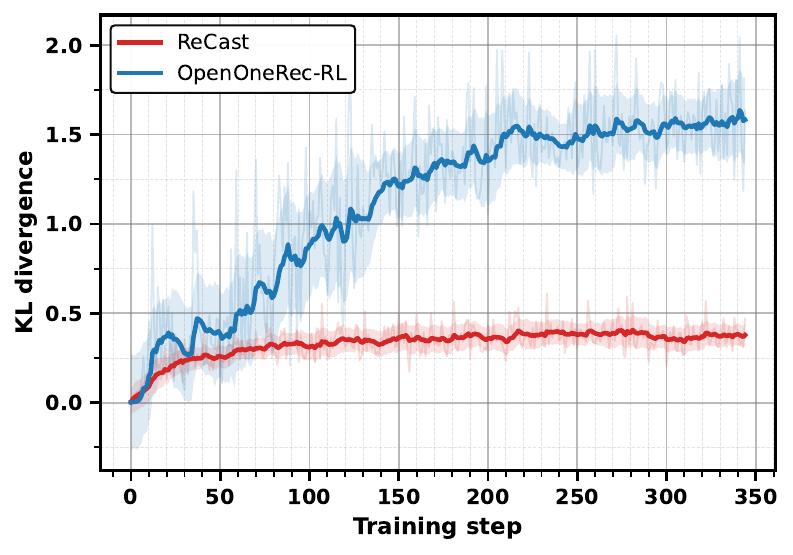}
        \caption{KL loss.}
        \label{fig:recast_stability_kl}
    \end{subfigure}
    \hfill
    \begin{subfigure}[t]{0.32\textwidth}
        \centering
        \includegraphics[width=\linewidth]{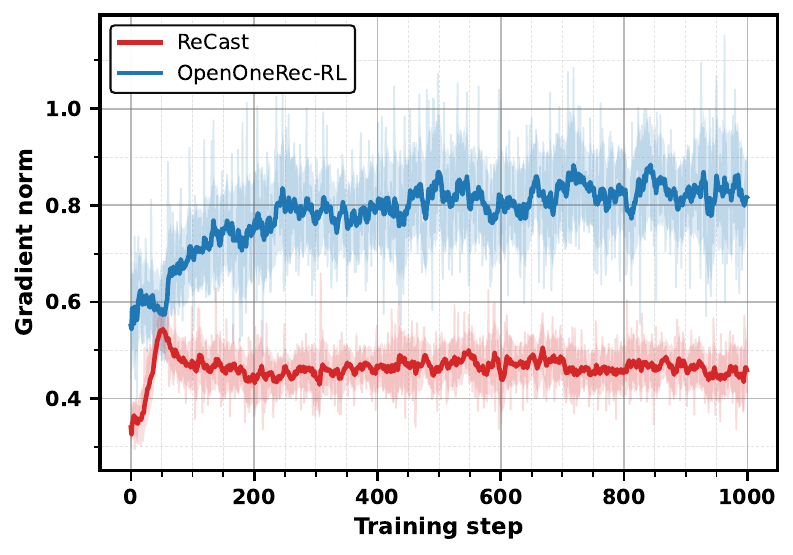}
        \caption{Gradient norm.}
        \label{fig:recast_stability_grad}
    \end{subfigure}
    \hfill
    \begin{subfigure}[t]{0.32\textwidth}
        \centering
        \includegraphics[width=\linewidth]{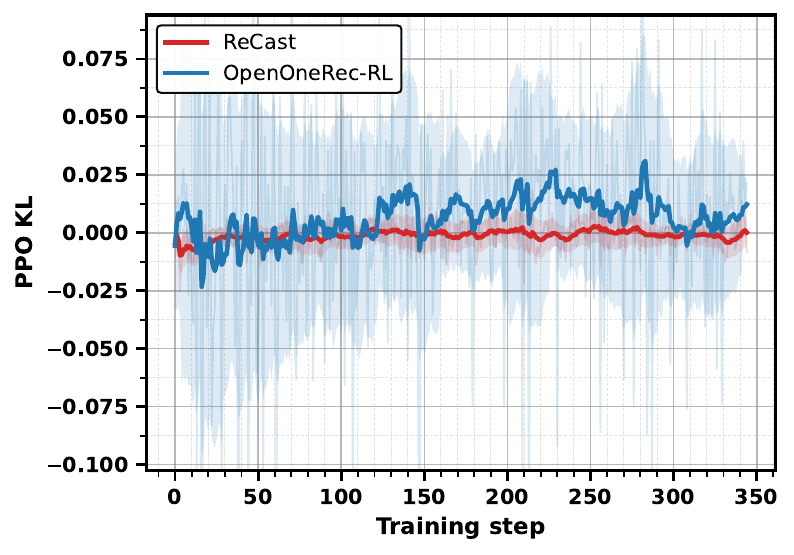}
        \caption{ppo-kl.}
        \label{fig:recast_stability_loss}
    \end{subfigure}
    \caption{Boundary-focused updating stabilizes optimization.}
    \label{fig:recast_stability}
\end{figure*}

Figure~\ref{fig:recast_stability} compares training dynamics between ReCast and OpenOneRec-RL using three standard indicators of update stability: \textbf{KL loss}, which tracks how strongly the current policy is pulled away from the reference policy; \textbf{PPO-KL}, which measures the effective step size of policy updates; and \textbf{gradient norm}, which reflects the magnitude and volatility of the optimization signal. Across all three metrics, ReCast exhibits smaller fluctuations, fewer spikes, and earlier stabilization.

This pattern is consistent with the signal path of ReCast. OpenOneRec-RL distributes credit through within-group reward statistics, so the update depends strongly on accidental group composition under sparse-hit supervision. ReCast instead concentrates learning on the local boundary defined by the strongest positive and the hardest negative. Policy updates therefore depend less on incidental group statistics and more on a stable boundary-supported signal, yielding smoother optimization.

\subsubsection{Ablation Study}

\begin{wrapfigure}{r}{0.5\linewidth}
    \centering
    \vspace{-0.6em}
    \includegraphics[width=\linewidth]{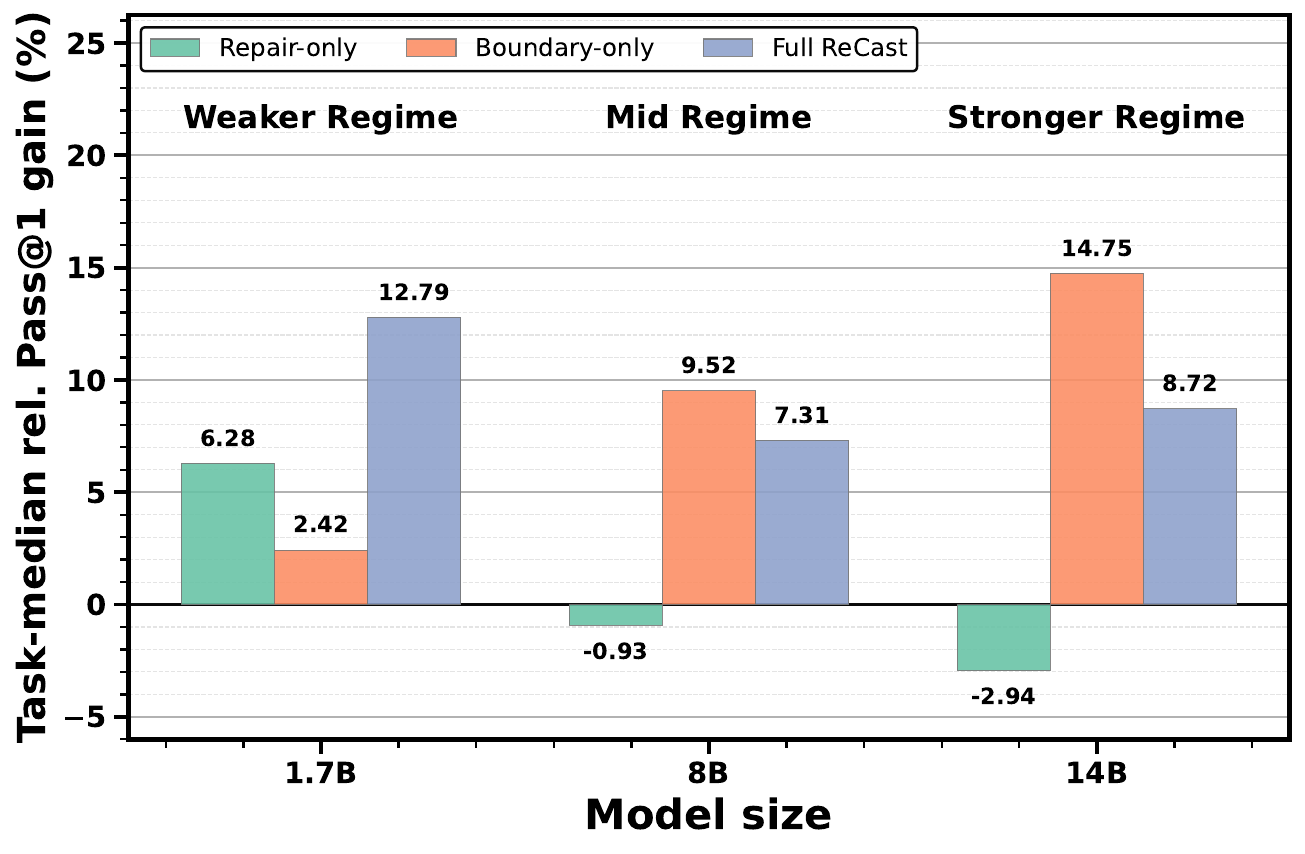}
    \vspace{-0.8em}
    \caption{Regime-dependent roles of repair and boundary-focused update. }
    \label{fig:ablation_regime_shift}
    \vspace{-0.8em}
\end{wrapfigure}

To separate the roles of the two components, we compare four variants: \textbf{OpenOneRec-RL}, \textbf{Repair-only}, \textbf{Boundary-only}, and \textbf{Full ReCast}. Figure~\ref{fig:ablation_regime_shift} reveals a clear regime-dependent pattern. In the weaker regime (\textbf{1.7B}), \textbf{Repair-only} contributes more than \textbf{Boundary-only}, indicating that the dominant bottleneck is still learnability itself: natural positives are scarce, and many sampled groups have not yet become trainable learning events.

As the regime becomes stronger (\textbf{8B} and \textbf{14B}), this relationship reverses sharply. \textbf{Boundary-only} becomes the stronger single-component variant, while \textbf{Repair-only} can turn mildly negative. This suggests that once the policy can already produce naturally trainable groups more reliably, the main difficulty is no longer making learning possible, but making existing learning signals more precise. In this setting, forced anchor injection may no longer align with the model's natural sampled boundary and can instead introduce bias into the update.

This ablation should therefore be read primarily as evidence of component roles, rather than as a claim that any fixed recipe remains uniformly optimal across all backbones. What it shows is a regime-dependent tradeoff within ReCast: repair is most useful when learnability is scarce, whereas boundary-focused updating carries more of the benefit once natural trainability improves.

\subsection{Scaling Under Search--Update Decoupling (for Q3)}

ReCast changes the scaling regime of group-based RL. Once actor-side learning no longer expands with the full rollout group, both larger backbones and broader rollout become more favorable under ReCast than under full-group updating.

\subsubsection{Scaling with Model Size}
\label{sec:scale_model_size}

We compare ReCast against OpenOneRec-RL at 1.7B, 8B, and 14B under the same RL comparison protocol. For each scale, we define the matched-performance budget ratio as
\[
\text{Budget Ratio}
=
\frac{\text{steps for ReCast to match OpenOneRec-RL@20K}}{20000}.
\]

\begin{figure*}[h]
    \centering
    \begin{subfigure}[t]{0.48\textwidth}
        \centering
        \includegraphics[width=\linewidth]{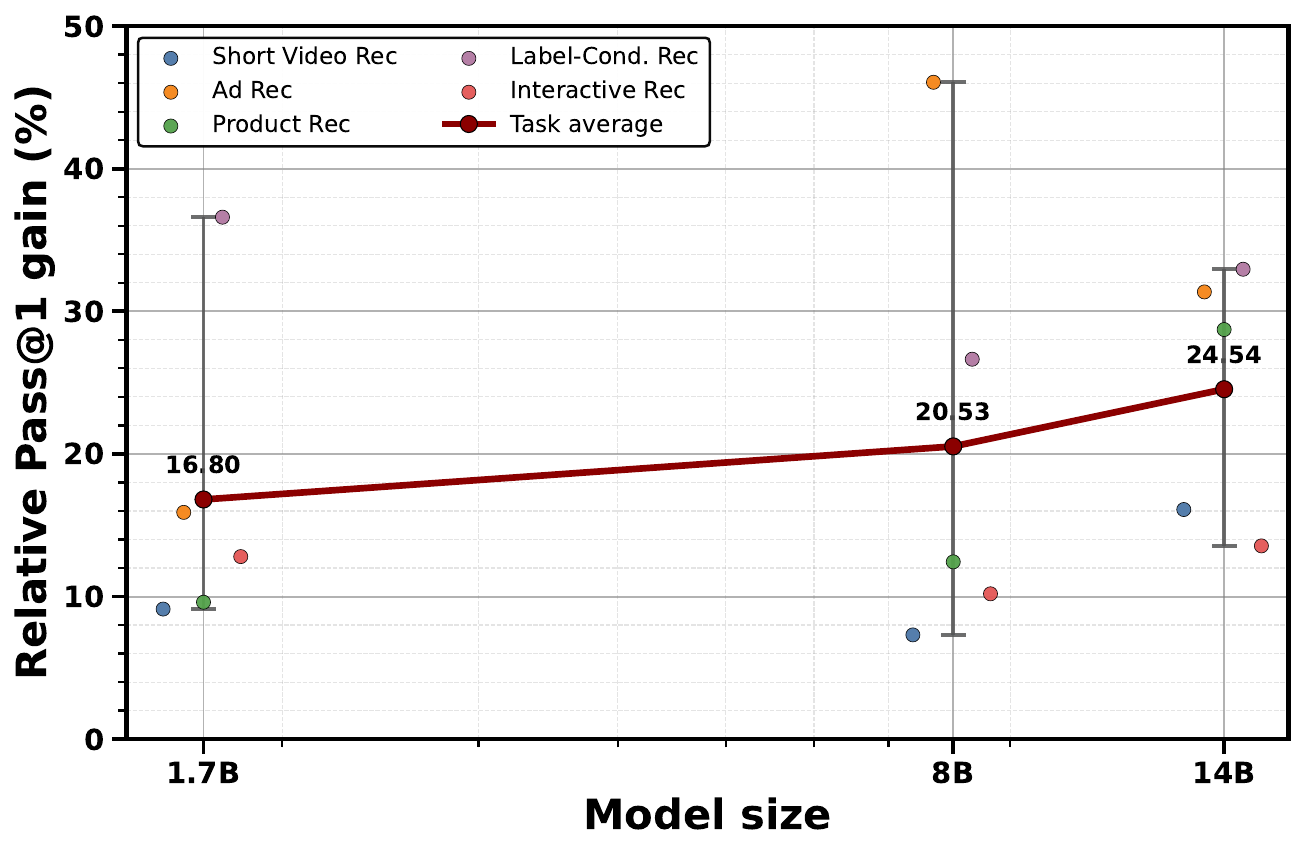}
        \caption{Relative Pass@1 gain vs. model size (log scale).}
        \label{fig:scaling_model_size_rel_gain}
    \end{subfigure}
    \hfill
    \begin{subfigure}[t]{0.48\textwidth}
        \centering
        \includegraphics[width=\linewidth]{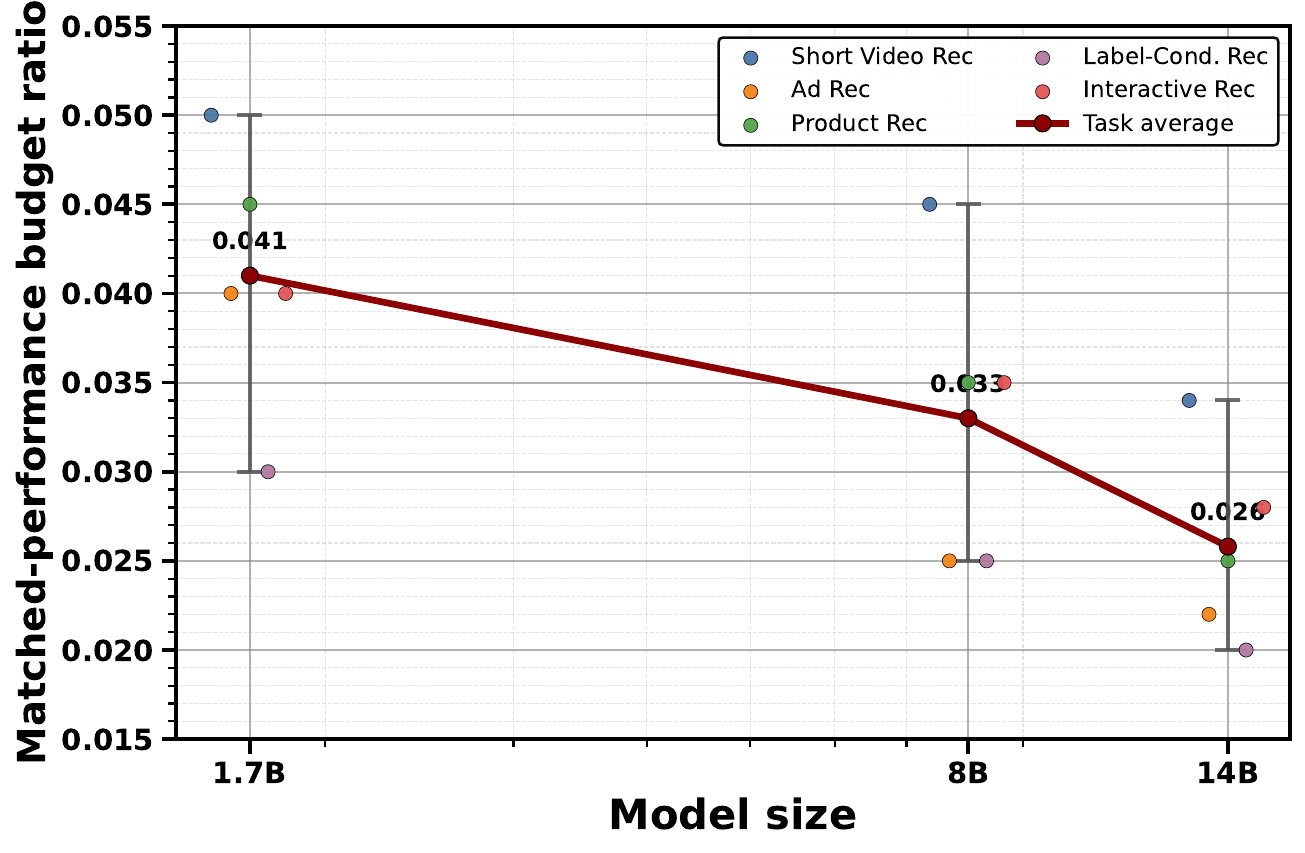}
        \caption{Matched-performance budget ratio vs. model size (log scale).}
        \label{fig:scaling_model_size_budget_ratio}
    \end{subfigure}
    \caption{Scaling with model size. Thin lines denote individual tasks and thick line denotes the average.}
    \label{fig:scaling_model_size}
    \vspace{-0.8em}
\end{figure*}

Figure~\ref{fig:scaling_model_size} shows a clear scaling trend. As model size increases, ReCast yields both larger fixed-budget gains and smaller matched-performance budget ratios. Averaged across tasks, the budget ratio decreases from \textbf{0.041} at 1.7B to \textbf{0.033} at 8B and further to \textbf{0.026} at 14B. In other words, ReCast needs only about \textbf{4.1\%}, \textbf{3.3\%}, and \textbf{2.6\%} of the 20K-step baseline budget at these three scales to reach the same target performance. Larger models therefore strengthen, rather than weaken, the case for ReCast. As actor-side optimization becomes more expensive with scale, keeping update support constant becomes increasingly valuable.

\subsubsection{Scaling with Search Width}
\label{sec:scale_search_width}

\begin{figure*}[t]
    \centering
    \begin{subfigure}[t]{0.48\textwidth}
        \centering
        \includegraphics[width=\linewidth]{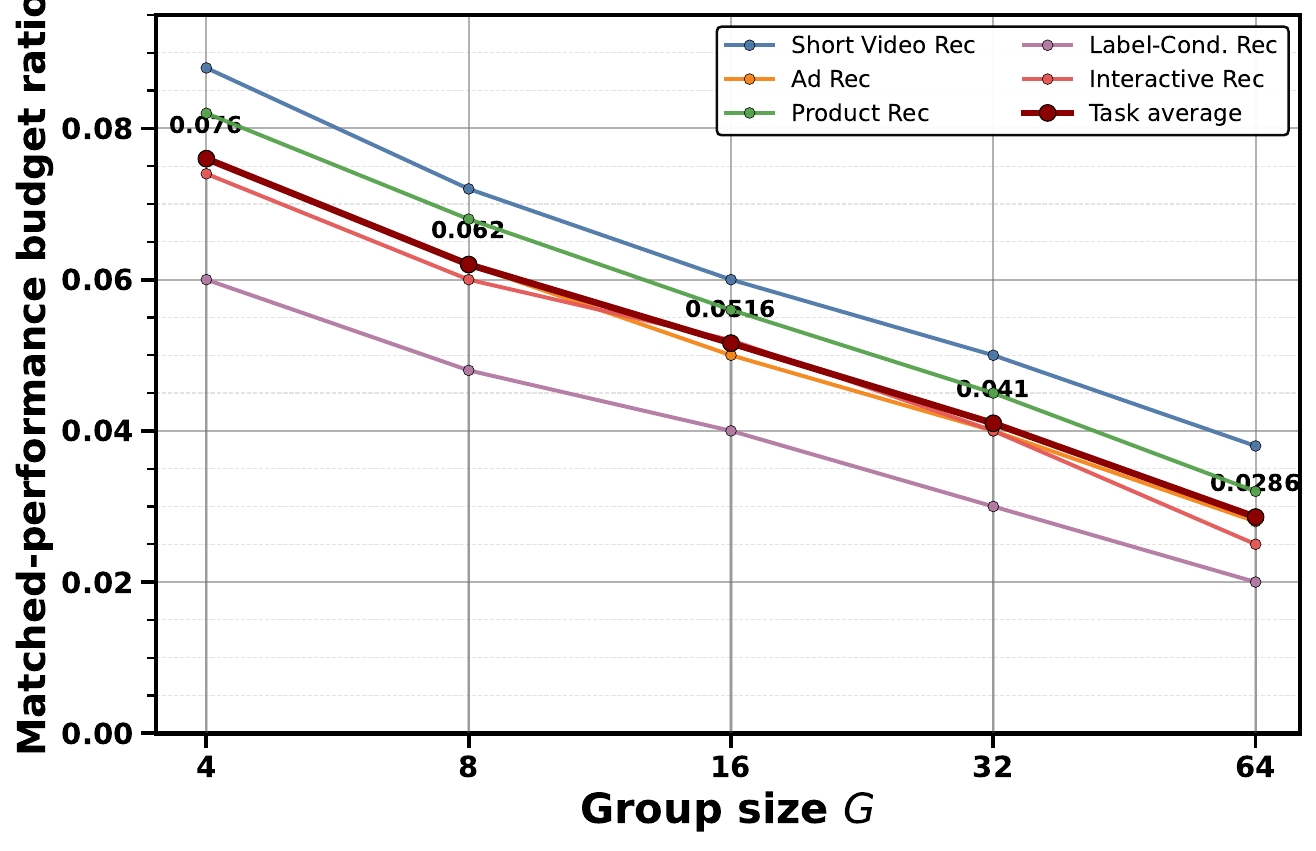}
        \caption{Matched-performance budget ratio vs. group size $G$.}
        \label{fig:scaling_group_size_budget_ratio}
    \end{subfigure}
    \hfill
    \begin{subfigure}[t]{0.48\textwidth}
        \centering
        \includegraphics[width=\linewidth]{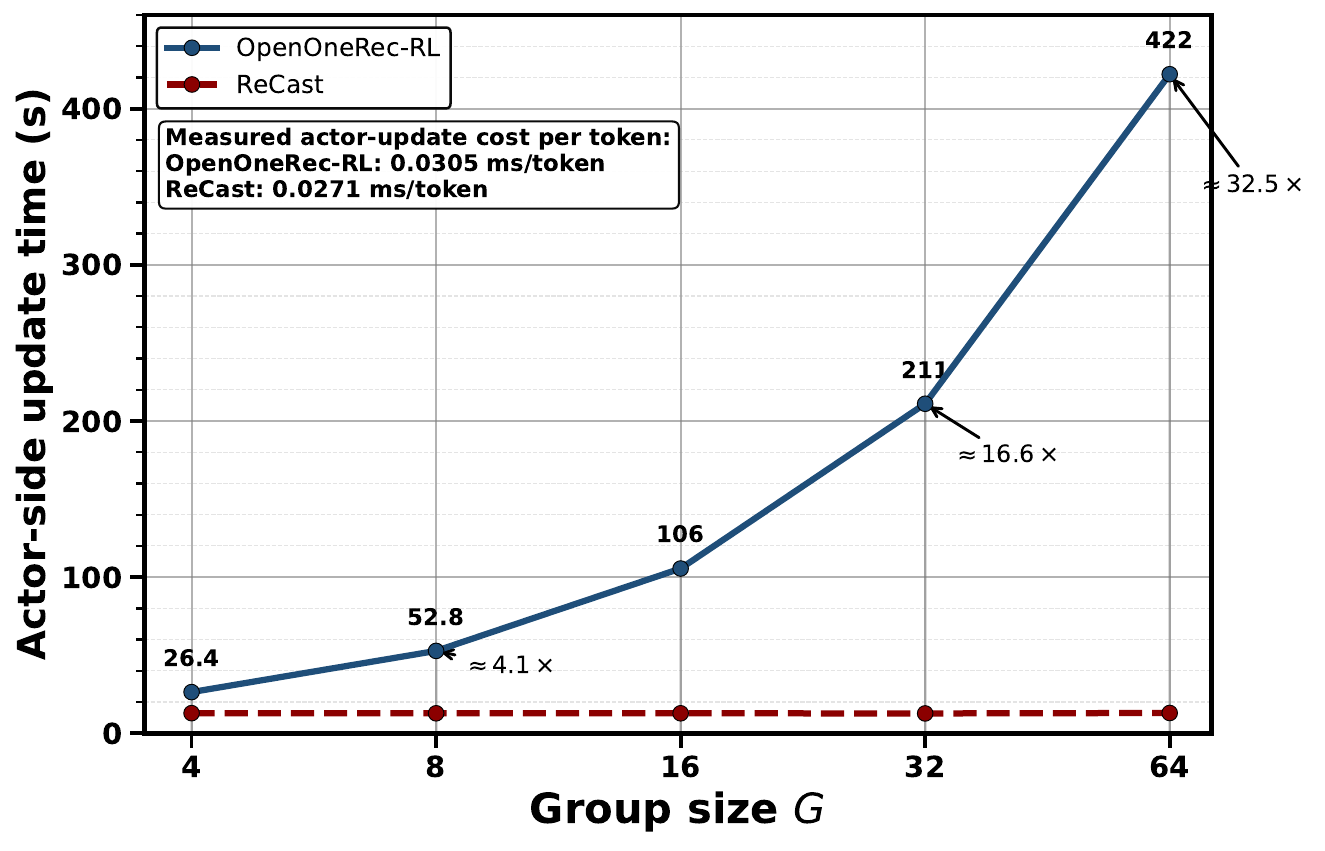}
        \caption{Normalized actor-side update cost vs. group size $G$.}
        \label{fig:scaling_group_size_actor_cost}
    \end{subfigure}
    \caption{Scaling with search width at fixed model size. }
    \label{fig:scaling_group_size}
\end{figure*}

We next fix the training setting and vary the rollout group size \(G\). Figure~\ref{fig:scaling_group_size} shows a clear divergence between the two methods. As \(G\) increases, ReCast achieves smaller matched-performance budget ratios, while its actor-side update cost grows much more slowly than that of OpenOneRec-RL. At \(G{=}32\), for example, ReCast reduces actor-side update time from \textbf{211.0s} to \textbf{12.7s}, corresponding to a \textbf{16.6$\times$} speedup.

This difference follows directly from search--update decoupling. Under full-group updating, increasing \(G\) broadens search but also expands actor-side learning in the same proportion. Under ReCast, increasing \(G\) still improves candidate coverage, but actor-side learning remains supported by a constant-size local boundary. Search width therefore becomes a more favorable scaling knob: broader rollout improves exploration without forcing proportionally broader actor updates.

\subsection{System-Level Efficiency}
\label{sec:efficiency}

The benefit of ReCast also appears directly in the training system. Table~\ref{tab:recast_efficiency} reports end-to-end results under a fixed setting with \(G{=}32\).

\begin{itemize}[leftmargin=*,noitemsep,topsep=2pt]
    \item \textbf{4.82$\times$ faster step time:} \textbf{371.54s} \(\rightarrow\) \textbf{77.00s}
    \item \textbf{16.60$\times$ faster actor update:} \textbf{211.04s} \(\rightarrow\) \textbf{12.71s}
    \item \textbf{16.5\% lower peak allocated memory:} \textbf{43.89 GB} \(\rightarrow\) \textbf{36.65 GB}
    \item \textbf{14.2\% higher actor MFU:} \textbf{0.0763} \(\rightarrow\) \textbf{0.0872}
\end{itemize}

\begin{table}[h]
\centering
\small
\setlength{\tabcolsep}{5pt}
\renewcommand{\arraystretch}{1.08}
\caption{System-level efficiency at fixed $G{=}32$.}
\label{tab:recast_efficiency}
\begin{tabular}{lccc}
\toprule
\textbf{Metric} & \textbf{OpenOneRec-RL} & \textbf{ReCast} & \textbf{Improvement} \\
\midrule
Step time (\texttt{timing\_s/step})         & 371.54 s & \textbf{77.00 s} & \textbf{4.82$\times$ faster} \\
Actor update (\texttt{update\_actor\_s})    & 211.04 s & \textbf{12.71 s} & \textbf{16.60$\times$ faster} \\
Old log-prob (\texttt{old\_log\_prob\_s})   & 56.27 s  & \textbf{4.13 s}  & \textbf{13.63$\times$ faster} \\
Reference forward (\texttt{ref\_s})         & 47.63 s  & \textbf{3.12 s}  & \textbf{15.25$\times$ faster} \\
Actor-side effective tokens / step          & 6.93M    & \textbf{0.47M}   & \textbf{14.78$\times$ fewer} \\
Peak allocated memory                       & 43.89 GB & \textbf{36.65 GB} & \textbf{16.5\% lower} \\
Actor MFU                                   & 0.0763   & \textbf{0.0872}  & \textbf{+14.2\%} \\
Actor update cost                           & 0.0305 ms/token & \textbf{0.0271 ms/token} & \textbf{11.0\% lower} \\
\bottomrule
\end{tabular}
\end{table}

These gains are not confined to a single stage. In addition to the large reduction in actor update time, \texttt{old\_log\_prob} and reference forward are accelerated by \textbf{13.63$\times$} and \textbf{15.25$\times$}, respectively. The benefit of ReCast therefore comes from shrinking actor-side active support throughout the entire pipeline.

The source of the gain is equally clear. ReCast reduces actor-side effective tokens per step from \textbf{6.93M} to \textbf{0.47M}, while the actor update cost per token does not increase and instead drops slightly from \textbf{0.0305 ms/token} to \textbf{0.0271 ms/token}. Its wall-clock improvement therefore comes primarily from reducing actor-side active workload. At the same time, memory usage decreases and actor MFU improves, showing that the gain is reflected not only in speed, but also in memory efficiency and hardware utilization.

\section{Discussion and Limitations}
\label{sec:discussion}

\subsection{Recommendation RL Exhibits a Regime Shift}

\paragraph{Optimization events are scarce.}
Sparse-hit generative recommendation is not generic preference alignment with a different reward. In many training steps, the sampled rollout is not yet a meaningful optimization event. The challenge is therefore not only how to optimize a learning signal, but when such a signal exists at all.

\paragraph{The bottleneck shifts with regime.}
Our results suggest that this bottleneck changes across regimes. In weaker regimes, the main problem is learnability: many sampled groups never become trainable events. In stronger regimes, sampled groups are more often naturally learnable, and the bottleneck shifts from making learning possible to preserving and refining already usable signals.

\paragraph{Efficiency can amplify mismatch.}
This also explains why more efficient RL updates do not always improve outcomes. ReCast increases effective gradient density, but efficiency amplifies whatever direction the gradient points. When learnability is the main bottleneck, this yields direct gains. Once a fixed sparse-reward RL objective drifts away from the backbone's structured decision boundary, however, more efficient updates can magnify that mismatch faster. The negative turn of repair in stronger backbones is one sign of this effect: once the model already forms trainable boundaries naturally, externally injected anchors may no longer align with them and can instead introduce bias.

\paragraph{A regime-dependent view of recommendation RL.}
The broader lesson is not that repair should always be used, but that recommendation-aware RL is not a fixed recipe. It is a regime-dependent problem: weaker regimes require explicit learnability restoration, whereas stronger regimes increasingly require signal refinement without destroying backbone structure.

\subsection{Limitations}

\begin{itemize}[leftmargin=*,itemsep=0.25em,topsep=0.2em]
    \item \textbf{Scope of validation.} Our study is limited to offline post-training for single-target next-item recommendation. This regime-shift view has so far been established only in this relatively clean sparse-hit setting. In longer-horizon, multi-objective, or delayed-feedback settings, the challenge will be not only to determine whether a sampled unit is learnable, but also when intervention begins to distort the model's natural learning boundary.

    \item \textbf{Unmeasured structural erosion.} We identify the phenomenon and its likely mechanism, but do not yet directly measure which SFT-induced structures are being eroded under stronger backbones. Repair can turn negative, and more efficient updates can amplify mismatch, but we do not yet know whether the main losses come from template structure, SID semantic alignment, candidate-constrained calibration, or broader representation drift.

    \item \textbf{Static repair within a fixed paradigm.} Repair remains static and rule-based. While useful in weaker regimes, it can become unnecessary or mildly harmful once the backbone already produces trainable groups naturally. This points to a more important future direction than a better fixed rule: adaptive repair, and more broadly, an adaptive RL-to-SFT interface that decides when to restore learnability, when to weaken intervention, and when to shift toward structure preservation.

    \item \textbf{The narrow reward interface beyond repair.} More broadly, the deeper issue may not be the current repair rule alone, but the sparse-scalar-reward interface itself. Strong SFT backbones carry structured behaviors along many dimensions, yet RL sees only the projection exposed by a single scalar reward. A more radical direction is therefore to move beyond single-scalar reward optimization toward denser multi-event targets, for example through world-model-based or multi-head event prediction that makes richer behavioral dimensions directly visible to RL optimization.
\end{itemize}

We therefore do not regard the current repair rule or boundary definition as final. They are better viewed as one concrete instance of a broader regime-aware signal-design perspective. The deeper open question is whether recommendation RL can learn, as the backbone and policy mature, not only when to transition from restoring learning events to preserving and refining usable ones, but also when a richer optimization interface is needed altogether.

\section{Related Work}
\label{sec:related}

Driven by rapid advances in frontier large language models \citep{gpt5,gemini3,qwen3.5,deepseekai2025deepseekr1incentivizingreasoningcapability,anthropic2026claudeopus46,liu2025advances}, generative recommendation has rapidly evolved into an increasingly capable LLM-based paradigm. It has progressed from early text-centric and instruction-following formulations \citep{p5,cui2022m6recgenerativepretrainedlanguage,tallrec,llara} to structured generation over semantic or discrete item IDs \citep{rajput2023recommender,10597986,letter,eager,eagerllm,unger,colarec,zhou2025onerec,liu2025onerecthinkintextreasoninggenerative}. More recent work has further strengthened this paradigm through improved item--text alignment and mixed-domain pre-training \citep{zhou2025openonerec,hao2025oxygenrec}, stronger SID--language grounding during supervised fine-tuning \citep{10597986,feng2026finegrainedsemanticsintegrationlarge,he2026reasoningsemanticidsenhances}, and post-training objectives for exploration, stability, and value-aware learning under sparse feedback \citep{shao2024deepseekmath,zheng2025groupsequencepolicyoptimization,jiang2026spendsearchpaysvalueguided,zhang2026reinforced,wang2025gflowgr}. Within this broader landscape, ReCast focuses on a more specific bottleneck: before improving reward shaping or optimization stability, a sampled group must first be learnable at all.

\subsection{RL for Generative Recommendation}

Existing generative recommendation RL methods mostly optimize learning signals after sampled rollouts have already been accepted as learning objects. OpenOneRec-RL~\citep{zhou2025openonerec} adopts GRPO-style within-group updates for recommendation generation, while OxygenRec~\citep{hao2025oxygenrec}, ECPO~\citep{zhou2025onerec}, and GBPO~\citep{zhou2025onerec2} improve recommendation RL through denser reward shaping, more conservative clipping, or gradient reshaping.

More recent work expands this design space along other axes. \citep{xie2026sagesequenceleveladaptivegradient} studies sequence-level and multi-objective optimization under sparse and rejection-dominated feedback, while V-STAR~\citep{jiang2026spendsearchpaysvalueguided} improves decoding-time exploration and within-group credit assignment through value-guided search and sibling-relative comparison. A related line introduces stronger reasoning or alignment machinery into recommendation RL, for example through reasoning-oriented optimization over Semantic IDs~\citep{he2026reasoningsemanticidsenhances} or joint reasoning-and-alignment training pipelines~\citep{hong2025generativereasoningrecommendationllms}. RISER~\citep{ding2026towards} is the closest in spirit to our work: it also identifies low sample utilization and instability in recommendation RL, but addresses them by converting failed rollouts into pairwise preference data together with additional diversification and token-level stabilization.
Overall, these methods improve reward shaping, objective stabilization, and even auxiliary supervision after rollout. By contrast, ReCast asks an earlier question: whether the sampled group is learnable in the first place.

\subsection{Alternative Training Paradigms and Semantic Foundations}

A separate line of work changes the supervision object or the semantic substrate rather than the within-group RL signal. GFlowGR~\citep{wang2025gflowgr} formulates generative recommendation as a multi-step generation problem and applies GFlowNets to provide value-aware supervision over item trajectories. TS-Rec~\citep{feng2026finegrainedsemanticsintegrationlarge} instead strengthens the representation substrate itself through semantic-aware SID initialization and token-level alignment. These methods are complementary to ReCast: they redesign what is supervised or how item tokens are represented, whereas ReCast studies sparse-hit group-based RL and focuses on restoring group-level learnability before refining the local positive--negative boundary.

\section{Conclusion}
\label{sec:con}

We identify a mismatch between generic group-based RL and sparse-hit generative recommendation: sampled groups are easy to obtain, but usable learning units are not. ReCast addresses this mismatch by restoring minimal learnability for all-zero groups and concentrating updates on the strongest local positive--negative boundary through a constant-size active subset. Across multiple generative recommendation tasks, ReCast consistently improves over the baseline, reaches useful learning substantially earlier, and exhibits a widening matched-budget advantage with model scale. More broadly, our results suggest that, for generative recommendation, the decisive RL problem is not only how to assign rewards, but how to construct learning signals from sparse, structured supervision.

\newpage
\bibliography{ref}
\bibliographystyle{abbrvnat}

\end{document}